\documentclass[cameraready]{Interspeech}
\usepackage{fontspec}
\usepackage{polyglossia}
\setmainlanguage{english}
\newfontfamily\ipafont[
  Path=./fonts/,
  BoldFont=CharisSIL-Bold.ttf,
  ItalicFont=CharisSIL-Italic.ttf,
  BoldItalicFont=CharisSIL-BoldItalic.ttf
]{CharisSIL-Regular.ttf} % full IPA coverage

\usepackage{microtype}        % improves spacing & reduces hyphenation
\usepackage{array}

\usepackage{booktabs}
\usepackage{multirow}
\usepackage{colortbl}
\usepackage{graphicx}
\usepackage{float} 
\usepackage{amsmath}
\usepackage[dvipsnames]{xcolor}
\usepackage{flafter} 
\usepackage{hyperref}

\definecolor{openai}{RGB}{200, 230, 245}    % cool slate blue
\definecolor{metaai}{RGB}{220, 235, 220}    % soft teal
\definecolor{omni}{RGB}{230, 220, 245}      % slightly deeper slate (sub-section)
\definecolor{ourmodels}{RGB}{240, 220, 200} % muted sage

\newfontfamily\ethiopicfont[
  Path=./fonts/,
  Script=Ethiopic
]{NotoSerifEthiopic-Regular.ttf}
\newcommand{\ethiopic}[1]{{\ethiopicfont #1}}
\newcommand{\ci}[2]{\textsubscript{#1}}

\title{Ethio-ASR: Joint Multilingual Speech Recognition and Language Identification for Ethiopian Languages}

\author{
Badr M. Abdullah$^{\heartsuit}$,
Israel Abebe Azime$^{\heartsuit}$,
Atnafu Lambebo Tonja$^{\dagger}$,
Jesujoba O. Alabi$ ^{\heartsuit}$, \\
Abel Mulat Alemu$^{\ddagger}$,
Eyob G. Hagos$^{\S}$,
Bontu Fufa Balcha$^{\spadesuit}$,
Mulubrhan A. Nerea$^{\clubsuit}$, \\
Debela Desalegn Yadeta$^{\spadesuit}$,
Dagnachew Mekonnen Marilign$^{\S}$,
Amanuel Temesgen Fentahun$^{\spadesuit}$, \\
Tadesse Kebede$^{\triangle}$,
Israel D. Gebru$^{\Diamond}$,
Michael Melese Woldeyohannis$^{\spadesuit}$, \\
Walelign Tewabe Sewunetie$^{\star}$,
Bernd Möbius$^{\heartsuit}$,
Dietrich Klakow$^{\heartsuit}$
}

\address{
$^{\heartsuit}$Saarland University, Germany \quad $^{\dagger}$University College London, UK,  \\ $^{\ddagger}$Ethiopian AI Institute, Ethiopia  \quad $^{\S}$HiLCoE, Ethiopia \quad $^{\spadesuit}$Addis Ababa University, Ethiopia, \\
$^{\clubsuit}$University West, Sweden \quad $^{\triangle}$Haramaya University, Ethiopia \quad $^{\Diamond}$Ethiopic.ai\\
$^{\star}$AIMS - Research and Innovation Centre, Rwanda
}

\email{badr.nlp@gmail.com}

\keywords{multilingual speech recognition, language identification, Ethiopian languages}

\usepackage{comment}

\begin{document}

\maketitle

\begin{abstract}
We present Ethio-ASR, a suite of multilingual CTC-based automatic speech recognition (ASR) models jointly trained on five Ethiopian languages: Amharic, Tigrinya, Oromo, Sidaama, and Wolaytta. These languages belong to the Semitic, Cushitic, and Omotic branches of the Afroasiatic family, and remain severely underrepresented in speech technology despite being spoken by the vast majority of Ethiopia's population. We train our models on the recently released WAXAL corpus using several pre-trained speech encoders and evaluate against strong multilingual baselines, including OmniASR. Our best model achieves an average WER of 30.48\% on the WAXAL test set, outperforming the best OmniASR model with substantially fewer parameters. We further provide a comprehensive analysis of gender bias, the contribution of vowel length and consonant gemination to ASR errors, and the training dynamics of multilingual CTC models. Our models and codebase are publicly available to the research community.
\end{abstract}

\section{Introduction}

Ethiopia is one of the African countries characterized by extraordinary cultural and linguistic diversity. 
According to the Ethnologue linguistic database, the country is home to  \textasciitilde128 million people who collectively speak more than 87 living indigenous languages \cite{ethnologue2025}. 
Nevertheless, the vast majority of Ethiopians remain excluded from the rapid adoption of voice-based technologies, as existing automatic speech recognition (ASR) systems provide little or no support for Ethiopian languages. 
This disparity is not merely a technological gap but a form of digital exclusion: speakers of languages unsupported by voice technology are effectively locked out of the growing ecosystem of voice-enabled AI services~\cite{10.1016/j.specom.2013.07.008}.
%rom healthcare and education to e-governance and accessibility tools. 
As speech technology continues to advance, the lack of support for low-resource languages risks widening existing inequalities rather than closing them~\cite{imam-etal-2025-automatic, diack2026waxal, pratap2024mms, alabi-etal-2025-charting}.

% add figure 1 
\begin{figure}[ht]
    \centering
    \includegraphics[width=\linewidth]{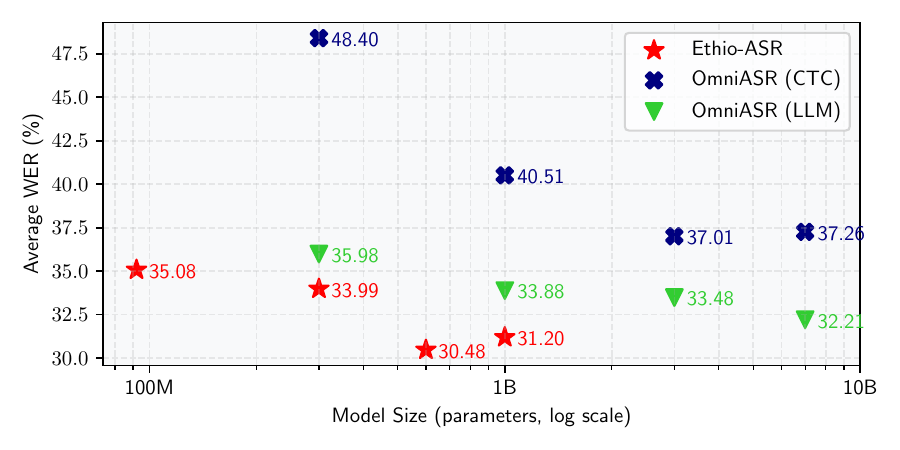}
\vspace{-16pt}
    \caption{Average WER (\%) versus model size, where lower WER  indicates better performance. Our Ethio-ASR models consistently achieve lower WER across all sizes, outperforming CTC and LLM-based OmniASR baselines.}
    \label{fig:figure-1}
\vspace{-16pt}
\end{figure}

\begin{table*}[ht]
\centering
\footnotesize
\caption{UDHR Article~1 across five Ethiopian languages in this paper with English translation as reference.}
\vspace{-6pt}
\label{tab:udhr}
\renewcommand{\arraystretch}{1.1}
\begin{tabular}{p{1.2cm}>{\centering\arraybackslash}p{1.0cm}p{11cm}p{1.2cm}}
\toprule
\rowcolor{gray!15}
\textbf{Language} & \textbf{ISO} & \textbf{Text sample} & \textbf{Script}\\
\midrule
English & \texttt{ENG}
  & All human beings are born free and equal in dignity and rights. & Latin \\
Amharic & \texttt{AMH}
  & \ethiopic{የሰው ልጆች ሁሉ ነጻ፥ በክብርም በመብትም እኩል ሆነው ተወልደዋል።}  & Ethiopic \\
Oromo & \texttt{ORM}
  & Namooti hundinuu birmaduu ta'anii mirgaa fi ulfinaanis wal-qixxee ta'anii dhalatan. & Latin\\
Tigrinya & \texttt{TIR}
  & \ethiopic{ብመንፅር  ክብርን  መሰልን  ኩሎም ሰባት እንትውለዱ ነፃን ማዕሪን እዮም።}  & Ethiopic \\
Sidaama &  \texttt{SID}
  & Manchi beetti kalaqamunni wolaphinoho. Ayirrinyunninna qoossotennino taaloho.  & Latin\\
Wolaitta & \texttt{WAL}
  & Ubba asaykka la'an daanawu yelettiis, qassikka bonchchuwaaninne maatan lagge. & Latin
  \\
\bottomrule
\end{tabular}
\vspace{-12pt}
\end{table*}

In this work, we focus on five Ethiopian languages, collectively spoken by the vast majority of Ethiopia's population: Amharic, Oromo, Tigrinya, Sidaama, and Wolaytta. 
These languages belong to the Afroasiatic language family and present a range of challenges for ASR systems, including rare phonological contrasts and complex morphology. While prior work has developed ASR systems for some of these languages (see Section~\ref{sec:related} for a comprehensive survey), these efforts have focused predominantly on Amharic, and crucially, no open-access models have been released. 
Large-scale multilingual models such as Whisper~\cite{radford2023robust} and SeamlessM4T~\cite{barrault2023seamless} offer little or no support for Ethiopian languages, as their performance remains far below practical utility. 
Taken together, these factors leave a critical gap: there are currently no publicly available, high-quality ASR models covering Ethiopian languages that can serve as a foundation for downstream NLP applications.
To address this gap, we make the following contributions:
\begin{itemize}
    \item We present Ethio-ASR, a suite of multilingual CTC-based ASR models that jointly perform ASR and language identification (LID) across five Ethiopian languages, trained on the largest transcribed Ethiopian speech corpus to date (Sections \ref{sec:data} and \ref{sec:model}).
    \item We benchmark our models against strong baselines including the recent OmniASR models, demonstrating that our models outperform all existing systems while using a fraction of the parameter count and inference cost (Section \ref{sec:experiments}).
    \item We provide a comprehensive analysis covering gender fairness, the effect of two linguistic features on ASR errors, and the training dynamics of our multilingual models (Section \ref{sec:analysis}).
    \item We will release our models\footnote{\href{https://huggingface.co/collections/badrex/ethio-asr}{https://huggingface.co/collections/badrex/ethio-asr}} and codebase\footnote{\href{https://github.com/badrex/Ethio-ASR}{https://github.com/badrex/Ethio-ASR}} to support future research and community-driven development of Ethiopian speech technology.
\end{itemize}

\section{Ethiopian Languages: An Overview}

% Ethiopia is one of the African countries characterized by extraordinary cultural and linguistic diversity. According to the Ethnologue linguistic database, the country is home to \textasciitilde128 million people who collectively speak more than 87 living indigenous languages \cite{ethnologue2025}. 
Most Ethiopians speak Afroasiatic languages belonging to the Semitic or Cushitic branches, while minority communities speak Nilo-Saharan languages \cite{bender1972language}. 
Although Amharic historically functioned as the sole official language and medium of instruction, Afar, Oromo, Somali, and Tigrinya have recently been recognized as official working languages.
In this section, we introduce the five Ethiopian languages in our research, with particular focus on their phonological features and writing systems.

\subsection{Amharic}

Amharic is an Ethio-Semitic language, a subgroup within the Semitic branch of Afroasiatic language family. 
It is spoken as a mother tongue by \textasciitilde29.3\% of the population and is the most learned second language across Ethiopia.
Phonologically, the Amharic phoneme inventory is characterized by several consonant sounds absent from English. Concretely, it features ejective consonants: speech sounds produced with a glottalic egressive airstream ({\ipafont /pʼ, tʼ, kʼ, kʷʼ, t͡ʃʼ, sʼ/}). The phoneme inventory also includes the glottal stop {\ipafont /ʔ/}, the palatal nasal {\ipafont /ɲ/}, and labialized velars ({\ipafont /kʷ, ɡʷ/}).
% Phonologically, Amharic is characterized by several consonant sounds that are absent in English including ejective stops and affricates ({\ipafont /pʼ, tʼ, kʼ, kʷʼ, t͡ʃʼ, sʼ/}), articulated with a glottalic egressive airstream, in addition to the glottal stop {\ipafont /ʔ/}, the palatal nasal {\ipafont /ɲ/}, labialized velars ({\ipafont /kʷ, ɡʷ/}), and an allophonic voiced bilabial fricative {\ipafont [β̞]}, whereby the phoneme {\ipafont /b/} is often realized as an approximant in intervocalic positions. 
The vowel system consists of the vowels: ({\ipafont /i, e, a, o, u, ə/}), along with two central vowels ({\ipafont /ɨ, ä/}), which are not present in English \cite{bender1972language, TACHBELIE2014181}. 
The syllable structure is relatively constrained since only a consonant-vowel (CV) sequence may occur in onset position, while initial consonant clusters (e.g., CCV) are disallowed. 
Word-final clusters are limited to two consonants (CVC or CVCC patterns). Amharic is written in the Ethiopic (Ge'ez) script.

\subsection{Oromo}

Oromo, or Afaan Oromoo, is a Cushitic language spoken by \textasciitilde34\% of Ethiopia's population, making it Ethiopia's most widely spoken first language. 
%and the fourth largest African language by number of native speakers.
Oromo's consonant inventory features two ejectives ({\ipafont /tʼ, kʼ/}) and an implosive {\ipafont /ɗ/}, none of which are present in English. Consonant gemination is contrastive and lexically distinctive, as illustrated by the minimal pair {\ipafont [samuu]} (`rot') versus {\ipafont [sammuu]} (`brain'). Oromo has five vowel qualities ({\ipafont /i, e, a, o, u/}) with phonemic length distinction, which yields minimal pairs such as {\ipafont [homaa]} (`nothing') versus {\ipafont [hoomaa]} (`mass of animals'). The permitted syllable structures are CV, CVC, and V; words may begin with a single consonant or a vowel, but never with a consonant cluster.
Since the early 1990s, Oromo has been written in a standardized Latin-based script known as Qubee \cite{bijiga2015development}.

% Oromo, or Afaan Oromoo, is a Cushitic language spoken by \textasciitilde 34\% of Ethiopia's population, which makes it  Ethiopia's most widely spoken first language and the fourth largest African language by number of native speakers.
% Oromo's consonant inventory features ejectives ({\ipafont /tʼ, kʼ/}) and an implosive {\ipafont /ɗ/}, none of which are present in English. 
% A salient phonological feature is gemination (consonant length), which is contrastive and lexically distinctive. 
% As a result, many words can be discriminated only by consonant length. 
% The minimal pair {\ipafont [samuu]} (`rot') versus {\ipafont [sammuu]} (`brain') illustrates meaning contrasts based solely on consonant length. 
% Oromo has five vowel qualities ({\ipafont /i, e, a, o, u/}) with phonemic length distinction, which yields minimal lexical pairs such as {\ipafont [homaa]} (`nothing') versus {\ipafont [hoomaa]} (`mass of people'). The syllable structures CV, CVC, and V are allowed. Words may begin with either a single consonant or a vowel, but never with a consonant cluster. Since the early 1990s, Oromo is written in a standardized Latin-based orthography known as Qubee \cite{bijiga2015development}.

\subsection{Tigrinya}
Tigrinya is an Ethio-Semitic language spoken primarily by communities in the Tigray Region of Ethiopia and in Eritrea.
Within Ethiopia, it ranks among the most widely spoken languages, with \textasciitilde6\% of the population speaking Tigrinya as a native language. 
While its vowel inventory closely resembles that of Amharic, Tigrinya exhibits a comparatively richer consonantal system. 
It features two pharyngeal consonants ({\ipafont /ħ, ʕ/}) and a uvular ejective fricative {\ipafont /xʼ/}, none of which are present in English or Amharic. 
As in Oromo, consonant gemination is phonemic in Tigrinya. Permitted syllable structures are CV and CVC \cite{ullendorff1955semitic, bender1972language}. 
Like Amharic, Tigrinya is written in the Ge'ez script.

% Tigrinya is an Ethio-Semitic language spoken primarily by communities in the Tigray Region of Ethiopia and in Eritrea. Within Ethiopia, it ranks among the most widely spoken languages with 6\% of Ethiopians speak Tigrinya as native speakers.
% While its vowel inventory closely resembles that of Amharic, Tigrinya exhibits a comparatively richer consonantal system that features two pharyngeal consonants ({\ipafont /ħ, \ipafont ʕ/}) and an uvular ejective fricative {\ipafont /xʼ/}, which are not present in English nor Amharic. 
% Tigrinya syllable structure permits CV and CVC patterns. 
% Like Amharic, it is written in the Ge'ez script.

\subsection{Sidaama}

Sidaama is a Cushitic language spoken in the Sidama Region of southern Ethiopia by \textasciitilde4\% of the population. Its vowel system consists of five qualities ({\ipafont /i, e, a, o, u/}), each can also occur in long forms ({\ipafont /iː, eː, aː, oː, uː/}). Vowel length is contrastive, as illustrated by the minimal pair {\ipafont [sinna]} (`branches') versus {\ipafont [siinna]} (`coffee cups'), and consonant gemination is likewise phonemic. The glottal stop {\ipafont /ʔ/} functions as a phoneme. All Sidaama words end in vowels, and consonant clusters are limited to two consonants occurring only intervocalically across syllable boundaries \cite{kawachi2007grammar}. 
Since 1993, Sidaama has been written using a Latin-based script. 

% Sidaama is a Cushitic language spoken in the Sidama Region of southern Ethiopia by \textasciitilde4\% of the population. Its vowel system consists of five short vowels ({\ipafont /i, e, a, o, u/}) and their long counterparts ({\ipafont /i:, e:, a:, o:, u:/}). 
% Vowel length is contrastive, as illustrated by minimal pairs such as {\ipafont [sinna]} (`branches') versus {\ipafont [siinna]} (`coffee cups'). 
% Consonant gemination is likewise contrastive. 
% The glottal stop  {\ipafont /ʔ/} functions as a phoneme. 
% All Sidaama words end in vowels, and consonant clusters are limited to two consonants, occurring only intervocalically across syllable boundaries. 
% Since 1993, Sidaama is written using a Latin-based orthography.

\subsection{Wolaytta}

Wolaytta is an Omotic language spoken by \textasciitilde2.2\% of Ethiopia's population. Unlike the previously discussed Semitic (Amharic, Tigrinya) and Cushitic (Oromo, Sidaama) languages, Wolaytta belongs to the Omotic branch of Afroasiatic, though it shares several areal phonological features with them.
As in Oromo and Sidaama, the vowel system consists of five qualities occurring in both short and long forms, and both vowel length and consonant gemination are phonemic. Wolaytta shares ejective consonants with the other Ethiopian languages discussed; however, unlike Tigrinya, it lacks pharyngeal consonants ({\ipafont /ħ, ʕ/}) \cite{Wakasa2008descriptive}. 
% A feature distinguishing Wolaytta from the other languages introduced here is its use of lexical tone (e.g., high vs.\ low), which may mark grammatical or lexical contrasts but is not represented in the standard orthography.
Syllable structure generally conforms to CV and CVC patterns. 
Wolaytta is written in a standardized Latin-based script. 
% in which vowel length and consonant gemination are marked through letter doubling.

% Wolaytta is spoken by \textasciitilde2.2\% of the population of Ethiopia. 
% Unlike the previously discussed Semitic (Amharic, Tigrinya) and Cushitic (Oromo, Sidaama) languages, Wolaytta belongs to the Omotic branch of Afroasiatic, though it shares several areal phonological features with them. 
% As in Oromo and Sidaama, the vowel system consists of five qualities that occur in both short and long forms. 
% Therefore, vowel length distinction and consonant gemination are both phonemic. 
% Wolaytta also shares ejective consonants with the other Ethiopian languages discussed; however, unlike Tigrinya it lacks pharyngeal consonants ({\ipafont /ħ, ʕ/}).
% A feature that distinguishes Wolaytta from the other languages introduced here is its use of lexical tone (e.g., high vs.\ low), which may mark grammatical or lexical contrasts but is not represented in the standard orthography.
% Syllable structure generally conforms to simple CV and CVC patterns. 
% Wolaytta is written in a standardized Latin-based orthography.

\subsection{Writing Systems}

\textbf{Ethiopic Script (Ge\textquoteleft ez).} Amharic and Tigrinya are written in Ge\textquoteleft ez, a writing system in which each grapheme represents a consonant-vowel sequence (an abugida). The script consists of 33 basic consonantal symbols, each can systematically be modified across seven vowel orders \cite{meyer2016ethiopic, TACHBELIE2014181}. 
The first order ({\ipafont /Cä/}) represents the base form, from which the remaining consonant-vowel combinations are derived through consistent graphic modifications (e.g., \ethiopic{በ} {\ipafont /bä/} $\rightarrow$ \ethiopic{ቡ} {\ipafont /bu/}, \ethiopic{ቢ} {\ipafont /bi/}, \ethiopic{ባ} {\ipafont /ba/}, \ethiopic{ቤ} {\ipafont /be/}, \ethiopic{ብ} {\ipafont /bǝ/}, \ethiopic{ቦ} {\ipafont /bo/}). The preferred syllable structure in Ethio-Semitic languages is CV(C), and this phonotactic pattern is reflected in the script's design; word-initial consonant clusters are typically avoided and resolved through insertion of the vowel {\ipafont /ǝ/}. 
Consonant gemination is not marked orthographically  \cite{meyer2016ethiopic}. 
%, and vowel distinctions are indicated solely through modification of consonant base forms. 
The primary Ethiopic Unicode block contains 384 code points, which reflects the large grapheme inventory generated by the alphasyllabary  system.

\vspace{0.25cm}
\noindent
\textbf{Latin-based Script.} Oromo, Sidaama, and Wolaytta are written in standardized Latin-based scripts, although Ge\textquoteleft ez- and, in some contexts, Arabic-based scripts were historically used. These scripts adapt the Latin alphabet to represent phonemic contrasts absent from English through two principal strategies. First, Latin letters that are redundant or underutilized in English are reassigned to represent distinct phonemes. 
For example, the letters $\langle${\ipafont c}$\rangle$, $\langle${\ipafont x}$\rangle$, and $\langle${\ipafont q}$\rangle$ represent the ejective phonemes {\ipafont /t͡ʃʼ/, /tʼ/}, and {\ipafont /kʼ/}, respectively. Second, digraphs are used to represent phonemes absent from the standard Latin inventory; for instance, $\langle${\ipafont ph}$\rangle$ represents the ejective {\ipafont /pʼ/}, while $\langle${\ipafont dh}$\rangle$ represents the implosive {\ipafont /ɗ/}. The glottal stop {\ipafont /ʔ/} is  represented by an apostrophe $\langle${\ipafont ʼ}$\rangle$. Vowel length and consonant gemination are marked through letter doubling; e.g., Oromo {\ipafont [homaa]} (`nothing') versus {\ipafont [hoomaa]} (`mass of animals') for vowel length, and {\ipafont [samuu]} (`rot') versus {\ipafont [sammuu]} (`brain') for gemination.

\begin{table}[t]
\centering
\caption{Audio duration (hours) by language, split, and gender (M: male voices, F: female voices).}
\vspace{-6pt} 
\label{tab:data_stats}
\renewcommand{\arraystretch}{1.0}
\resizebox{\columnwidth}{!}{
\begin{tabular}{ll rrrrr}
\toprule
& & \texttt{AMH} & \texttt{ORM} & \texttt{TIR} & \texttt{SID} & \texttt{WAL} \\
\midrule
\rowcolor{gray!20}
\multicolumn{2}{l}{\color{Blue}\textbf{Train}} & & & & & \\
  & M   & 105.88 & 85.14  & 83.20  & 64.70  & 76.56  \\
  & F   & 83.83  & 104.75 & 98.67  & 127.53 & 120.75 \\
  & All & 189.71 & 189.89 & 181.87 & 192.23 & 197.32 \\
  \midrule
\rowcolor{gray!20}
\multicolumn{3}{l}{\color{Blue}\textbf{Validation}} & & & & \\
  & M   & 6.56   & 5.04   & 6.70   & 4.75   & 0.02   \\
  & F   & 7.41   & 9.42   & 10.09  & 10.59  & 9.58   \\
  & All & 13.97  & 14.46  & 16.79  & 15.34  & 9.60   \\
  \midrule
\rowcolor{gray!20}
\multicolumn{2}{l}{\color{Blue}\textbf{Test}} & & & & & \\
  & M   & 6.98   & 10.04  & 11.12  & 6.27   & 0.00   \\
  & F   & 9.26   & 8.20   & 9.14   & 11.53  & 12.40  \\
  & All & 16.24  & 18.24  & 20.26  & 17.80  & 12.40  \\
\bottomrule
\end{tabular}}
\vspace{-15pt}
\end{table}

\subsection{Challenges for ASR Systems}
\label{sec:asr_challenges}
% Ethiopian languages pose a range of challenges for ASR systems, which stem from underrepresentation in pre-training corpora, rare phonological contrasts, large grapheme inventories, and morphological complexity.

\textbf{Underrepresented phonemes.} Ejective consonants (e.g., {\ipafont /pʼ, tʼ, kʼ, t͡ʃʼ/}), pharyngeal consonants ({\ipafont /ħ, ʕ/}), and the implosive {\ipafont /ɗ/} are not common in high-resource languages used to pre-train multilingual foundation models such as Whisper \cite{radford2023robust} and XLS-R \cite{babu22_interspeech}. 
Therefore, the acoustic representations learned during pre-training are unlikely to effectively discriminate between these contrasts, and it is unknown whether fine-tuning on limited target language data would be sufficient to address their underrepresentation in the pre-training distribution.

\vspace{0.1cm}
\noindent
\textbf{Gemination and vowel length.} Consonant gemination and vowel length are lexically distinctive features and are orthographically marked through letter doubling in Oromo, Sidaama, and Wolaytta. 
Correctly capturing these length contrasts requires ASR systems to distinguish long from short segments under naturally variable speaking rates. 
Prior work has shown that vowel-length contrasts pose challenges for HMM-based ASR in Hausa and Wolof, and that explicitly modeling the contrast as distinct phoneme categories yields marginal yet consistent improvements \cite{gauthier16b_interspeech}. 
However, whether and to what extent current end-to-end ASR adaptation strategies can capture these contrasts remains an open question, which we investigate in Section~\ref{sec:length_analysis}.

\vspace{0.1cm}
\noindent
\textbf{Large grapheme inventory.} For Amharic and Tigrinya, the Ge’ez script features a large grapheme inventory through its consonant-vowel syllabary system. 
Grapheme frequency follows a long-tail distribution: a small number of high-frequency graphemes account for the majority of tokens in transcribed corpora, while a large proportion of the inventory has low frequency. 
In low-resource settings, training data is often too small to provide adequate coverage of all graphemes. 
%, which affects both acoustic and language modeling.

\vspace{0.1cm}
\noindent
\textbf{Morphological complexity.} All Ethiopian languages have rich morphological systems that yield large numbers of distinct wordforms through inflectional and derivational processes \cite{bender1972language, TACHBELIE2014181, gebre2010part, info14030195}. 
This substantially increases lexical diversity and out-of-vocabulary rates, posing challenges for both language modeling and ASR evaluation.

\section{Dataset}
\label{sec:data}

\subsection{WAXAL Dataset}

 % The dataset for this research comprises five Ethiopian languages, consisting of two Semitic (Amharic and Tigrinya), two Cushitic (Oromo and Sidama), and one Omotic (Wolaytta) language, which were extracted from a larger, publicly available corpus of 21 languages \cite{diack2026waxal}.

 The dataset for this research is the Ethiopian subset of the WAXAL corpus, a multilingual speech dataset for 21 African languages recently released under the CC-BY-SA-4.0 license \cite{diack2026waxal}. 
 The Ethiopian subset covers five Afroasiatic languages: two Ethio-Semitic (Amharic and Tigrinya), two Cushitic (Oromo and Sidaama), and one Omotic (Wolaytta). 
  Training data is balanced across languages, ranging from 181 to 197 hours per language, and the full dataset amounts to \textasciitilde1106 hours across all five languages and splits. 
  To our knowledge, this is the largest publicly available speech corpus for Ethiopian languages to date. 
  Table~\ref{tab:data_stats} summarises the audio duration by language, split, and gender, where gender metadata is self-identified by the speaker.

 % The dataset was collected using two primary modalities: scripted and unscripted. For the scripted modality, the process began with transcribing and validating random images from over 40 categories, which was then followed by the recording itself. The unscripted modality, however, comprised three distinct speech types: spontaneous, expert, and prompted speech described from a set of randomly selected images. The collection captured prosodic differences in each recording type, spanning free, emphatic, expressive, and question contexts. A key consistency across both modalities was that any utterance spoken in response to an image prompt had to directly relate to that image; otherwise, it was rejected.

 Speech data was collected using two modalities: scripted and unscripted. 
 In the scripted modality, text transcriptions of randomly selected images from over 40 categories were recorded by speakers. 
 The unscripted modality consists of three speech types: spontaneous, expert, and prompted speech, where speakers describe a set of randomly selected images. 
 Both modalities captured prosodic variation across four speech styles: free, emphatic, expressive, and interrogative (i.e., question context). 
 To ensure consistency across both modalities, any utterance spoken in response to an image prompt had to be directly related to the image; otherwise, it was rejected.

The collection workflow involved three distinct roles: data collectors, transcribers, and validators. 
Collectors contributed their voices by either reading transcriptions (scripted) or describing images (unscripted). 
Transcribers then transformed speech samples into text. 
Validators reviewed each recording for adherence to quality guidelines covering noise levels, silence duration, and transcription accuracy. 
Each recording/transcription was assessed by at least two validators, who verified the alignment among the image, audio, and transcript, regardless of the order in which they were produced.

% The corpus collection workflow was divided among three distinct groups: data collectors, transcribers, and validators. Collectors contributed audio by either reading transcriptions or describing images using free, emphatic, expressive, or questioning speech. Transcribers then documented the audio or the image descriptions. Finally, validators reviewed each recording for quality, ensuring adherence to guidelines on noise levels, silence length, and overall accuracy in addition to validating the transcription. A minimum of two validators assess every audio recording and transcription to verify the alignment between the image, the audio, and the transcript, without regard to the sequence of production.

\begin{figure}[t]
    \centering
    \includegraphics[width=\linewidth]{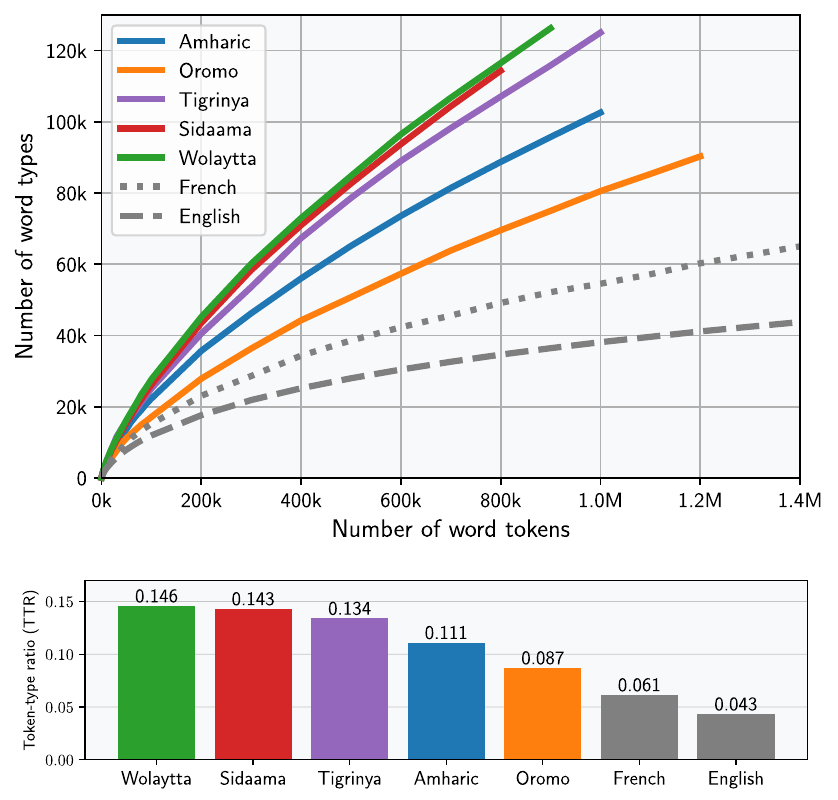}
    \vspace{-15pt} 
    \caption{Top: Vocabulary growth curves for Ethiopian languages (WAXAL) and English/French (Multilingual LibriSpeech) across corpus sizes up to 1.4M tokens. Bottom: Type-token ratio (TTR) at 800k tokens.}
    \label{fig:lexical_analysis}
    \vspace{-15pt}
\end{figure}

\subsection{Data-driven Lexical Analysis}

To empirically quantify the effect of morphological complexity,  we analyse vocabulary growth and type-token ratio (TTR) across the five Ethiopian languages in the WAXAL corpus and compare them against English and French from the Multilingual LibriSpeech corpus~\cite{panayotov2015librispeech, pratap20_interspeech} as high-resource reference points (Figure~\ref{fig:lexical_analysis}).  
Token-type ratio (TTR) has been shown to correlate with morphological complexity across languages \cite{kettunen2014can, juola1998measuring, schultz2006multilingual, bentz-etal-2016-comparison}. 
%Morphologically complex languages tend to have higher TTR values due to the larger number of distinct wordforms produced by inflectional and derivational processes. 
In Figure~\ref{fig:lexical_analysis}, we observe that all five Ethiopian languages exhibit substantially steeper type-token growth curves than English and French, indicating that new word types continue to emerge at a much higher rate as the corpus grows. 
This is also reflected in the TTR values, where Wolaytta (0.146), Sidaama (0.143), and Tigrinya (0.134) are more than three times higher than English (0.043). 
This analysis provides corpus-level evidence for the effect of morphological complexity discussed in Section~\ref{sec:asr_challenges}.

\section{Joint ASR and LID Modeling}
\label{sec:model} 
The five Ethiopian languages we study in this work share several phonological features and orthographic conventions within each script, which motivates a multilingual modeling approach. 
We adopt a joint LID and ASR setup in which both tasks are performed by a single shared CTC model, following established practices in multilingual end-to-end ASR \cite{watanabe2017language, bartelds2025ctc, peng2024owsm}.

\subsection{Model Architecture}

Our model consists of a speech encoder (initialized from a pre-trained checkpoint) and a linear projection to a shared output vocabulary. Transcription is performed using CTC decoding, where the output sequence for each utterance is defined as a language identification token prepended to the grapheme sequence as $\boldsymbol{y} = \langle \ \texttt{[LANG]}, \mathcal{G}_1, \mathcal{G}_2, \dots, \mathcal{G}_N \ \rangle$, where \texttt{[LANG]} $\in$ \{\texttt{[AMH]}, \texttt{[TIR]}, \texttt{[ORM]}, \texttt{[SID]}, \texttt{[WAL]}\} is the language token and $\mathcal{G}_1, \dots, \mathcal{G}_N$ is the grapheme sequence. The model is trained with the standard CTC loss over this target sequence, without a separate LID loss.

\subsection{Grapheme-based Vocabulary}

The output vocabulary is grapheme-based and covers both scripts used across the five languages. 
For Amharic and Tigrinya, the vocabulary includes 326 core Ge\textquoteleft ez graphemes, along with 29 Ethiopic punctuation marks and numerals. 
For Oromo, Sidaama, and Wolaytta, the vocabulary includes the 26 Latin letters and 25 Latin punctuation marks and numerals. 
Punctuation and numerals are retained for both scripts as they appear in the human transcriptions. 
The LID tokens are added as special symbols, giving a final vocabulary of 409 symbols in total including all special tokens.

\subsection{Training Objective}

Given an input speech $\boldsymbol{x}$ and its corresponding target sequence $\boldsymbol{y}$, the model is trained to maximize the CTC log-likelihood  $\mathcal{L} = -\log P_{\text{CTC}}(\boldsymbol{y} \mid \boldsymbol{x})$. 
The LID token is treated as an ordinary output symbol within the CTC framework, with no separate LID head. 
As a result, the model learns to predict LID jointly with transcription in a single pass, without requiring a loss weighting hyperparameter as in conventional multi-task setups. 
%Despite its simplicity, this design facilitates parameter sharing between the two tasks.

\section{Experiments}
\label{sec:experiments}

\begin{table*}[h!]
\centering
\caption{WER (\%, lower is better) on the test split of the WAXAL dataset. 
Bold indicates the best result per language among multilingual Ethio-ASR models, while underlined indicates best performance among all models.}
\vspace{-6pt} 
\label{tab:waxal_wer}
\resizebox{\textwidth}{!}{%
\begin{tabular}{lcc|ccccc|c}
\toprule
\textbf{Model} & \textbf{\# params} & \textbf{Decoder} & \textbf{Amharic} & \textbf{Tigrinya} & \textbf{Oromo} & \textbf{Wolaytta} & \textbf{Sidaama} & \textbf{Avg.} \\
\midrule

% openai section
\rowcolor{openai}
\multicolumn{9}{l}{\textbf{OpenAI ASR}} \\
\quad whisper-small     & 300M  & LM & 157.85 & 174.31 & 152.99 & 184.67 & 182.32 & 170.43 \\
\quad whisper-medium    & 786M  & LM & 195.05 & 198.06 & 192.64 & 247.30 & 232.91 & 213.19 \\
\quad whisper-large-v3  & 1.6B  & LM & 153.03 & 166.20 & 128.65 & 151.04 & 144.08 & 148.60 \\
\midrule

% meta + omni section
\rowcolor{metaai}
\multicolumn{9}{l}{\textbf{Meta ASR models}} \\
\quad seamless-m4t-v2-large  & 2B    & LM  & 103.75 & 100.00 & 100.00 & 100.00 & 100.00 & 100.75 \\
\quad mms-1b-all             & 1B    & CTC &  57.53 &  70.48 &  41.53 & 104.22 &  37.64 &  62.28 \\
\cmidrule{1-9}
\rowcolor{omni}
\multicolumn{9}{l}{\textbf{Meta OmniASR}} \\
%\multirow{4}{*}{\rotatebox{0}{CTC-based}}
\quad omniASR-ctc-300m-v2  & 300M & CTC & 49.15 & 58.11 & 40.77 & 52.90 & 41.08 & 48.40 \\
\quad omniASR-ctc-1b-v2  & 1B   & CTC & 37.44 & 50.15 & 31.34 & 46.35 & 37.26 & 40.51 \\
\quad omniASR-ctc-3b-v2  & 3B   & CTC & 32.41 & 45.91 & 27.91 & 43.44 & 35.38 & 37.01 \\
\quad omniASR-ctc-7b-v2  & 7B   & CTC & 32.48 & 46.21 & 27.79 & 44.58 & 35.21 & 37.26 \\
\cmidrule{1-9}
%\multirow{4}{*}{\rotatebox{0}{LLM-based}}
\quad omniASR-llm-300m-v2  & 300M & LLM & 30.95 & 46.10 & 27.33 & 41.43 & 34.10 & 35.98 \\
\quad omniASR-llm-1b-v2  & 1B   & LLM & 27.65 & 42.87 & 25.28 & 40.37 & 33.21 & 33.88 \\
\quad omniASR-llm-3b-v2  & 3B   & LLM &  26.83     &     42.32  &   24.80    &   40.36    &  32.91     &  33.48     \\
\quad omniASR-llm-7b-v2  & 7B   & LLM & 25.12 & 40.69 & \underline{23.59} & 39.22 & 32.46 & 32.21 \\
\midrule

% our models section
\rowcolor{ourmodels}
\multicolumn{9}{l}{\textbf{This work}} \\
\quad Ethio-ASR (\texttt{afrihubert}) & 94M    & CTC & 30.95 & 42.42 & 27.57 & 40.44 & 34.02 & 35.08 \\
\quad Ethio-ASR (mms-300)    & 300M   & CTC & 30.19 & 41.62 & 26.41 & 39.10 & 32.66 & 33.99 \\
\quad Ethio-ASR (\texttt{mms-1b})     & 1B  & CTC & 26.14 & 37.63 & \textbf{23.69} & \underline{\textbf{37.51}} & \textbf{31.02} & 31.20 \\
\quad Ethio-ASR (\texttt{w2v-bert-2.0})      & 600M   & CTC & \textbf{22.92} & \underline{\textbf{35.22}} & 24.44 & 38.19 & 31.65 & \textbf{30.48} \\
\cmidrule{1-9}
\quad monolingual SFT (\texttt{w2v-bert-2.0})     & 5$\times$600M & CTC & \underline{22.37} & 35.65 & 24.29 & 37.64 & \underline{30.04} & \underline{30.00} \\

\bottomrule
\end{tabular}%
}
\vspace{-12pt}
\end{table*}

\subsection{Experimental Setup}

\textbf{Baselines.} We evaluate our models against several strong multilingual baselines: the small, medium, and large encoder-decoder Whisper models \cite{radford2023robust}, the massively multilingual \texttt{mms-1b-all} model fine-tuned for ASR \cite{pratap2024mms}, the multimodal SeamlessM4T \cite{barrault2023seamless}, and the OmniASR models including both CTC- and LLM-based variants \cite{omnilingualasrteam2025omnilingualasropensourcemultilingual}. Of the baselines, Whisper has native support only for Amharic, \texttt{mms-1b-all} was trained on FLEURS dataset which covers Amharic and Oromo \cite{conneau2023fleurs}, and the OmniASR models were trained on a multilingual mixture that includes all five Ethiopian languages in this study.

\vspace{0.1cm}
\noindent
\textbf{Pre-trained Speech Encoders.} We experiment with four pre-trained self-supervised speech encoders and adapt them for ASR using multilingual supervised fine-tuning (SFT): AfriHuBERT~\cite{alabi25_interspeech}, a Transformer-based encoder pre-trained on African languages (94M parameters, CC-BY-NC-SA 4.0 license); MMS encoder~\cite{pratap2024mms},  pre-trained on over 1{,}000 languages in two sizes (300M and 1B parameters, CC-BY-NC 4.0 license); and wav2vec-BERT-2.0 \cite{barrault2023seamless}, a Conformer-based encoder pre-trained on 4.6M hours of multilingual speech (600M parameters, MIT license).

\vspace{0.1cm}
\noindent
\textbf{Training Hyperparameters.} All models are fine-tuned for seven epochs with an effective batch size of 32 samples, yielding \textasciitilde36.8k steps. We use the AdamW optimizer with a learning rate tuned over $\{3 \times 10^{-5}, 7 \times 10^{-5}, 3 \times 10^{-4}, 7 \times 10^{-4}\}$ and a linear warmup over the first 10\% of training steps. The convolutional feature extractor is frozen throughout training. Mixed precision training is used with \texttt{bfloat16}, except for AfriHuBERT where full \texttt{float32} precision is used. Models are evaluated on the validation split every 800 steps and the best checkpoint is selected based on $0.5 \times \text{WER} + 0.5 \times \text{CER}$. Our codebase is developed using the Hugging Face ecosystem and will be publicly released upon publication.

\subsection{Evaluation on WAXAL Dataset}

Although our models are trained on the complete Ethiopic character set and produce punctuations, we apply punctuation removal and normalize homophones in the Ge\textquoteleft ez script as post-processing before evaluation, following established best practices in the Ethiopic NLP community \cite{nigatu-etal-2025-case}.
Table~\ref{tab:waxal_wer} reports WER on the test split of the WAXAL dataset. 
% The results show that existing large-scale ASR models perform poorly on Ethiopian languages, while our approach using multilingual supervised fine-tuned (SFT) achieves substantial improvements across all five languages.

\vspace{0.1cm}
\noindent
\textbf{Baseline performance.} OpenAI Whisper and Meta Seamless M4T models show high WER across all languages, with several models exceeding 100\% WER. 
This trend indicates excessive insertions likely caused by the underrepresentation in the pre-training data or the lack of support for these languages (only Amharic was in the pretraining mixture of Whisper). 
The best-performing baseline among this group is \texttt{mms-1b-all} (avg.\ 62.28\%), which seems to benefit from multilingual fine-tuning on the FLEURS dataset which includes transcribed speech for Amharic and Oromo among many other African languages.

\vspace{0.1cm}
\noindent
\textbf{OmniASR.} The LLM-based OmniASR variants outperform their CTC-based counterparts, a gap that can be attributed to the autoregressive decoding strategy of the LLM-based models. 
However, it is important to note that OmniASR models were trained on a pre-released subset of the WAXAL corpus from July 2025, which means we cannot rule out overlap between their training data and the current WAXAL test split.

\vspace{0.1cm}
\noindent
\textbf{Ethio-ASR.} In Table~\ref{tab:waxal_wer}, one can observe that our multilingual models outperform all OmniASR variants while using significantly fewer parameters. 
Our best model, \texttt{Ethio-ASR (w2v-bert-2.0)} with 600M parameters, achieves an average WER of 30.48\%, outperforming the best OmniASR model overall, \texttt{omniASR-llm-7b-v2} (32.21\%), which has more than ten times the parameters. 
When considering only CTC-based models, even our smallest model, \texttt{Ethio-ASR (afrihubert)} with 94M parameters, outperforms all OmniASR CTC variants, the best of which is \texttt{omniASR-ctc-7b-v2} (37.26\%). 
This efficiency advantage has practical implications: OmniASR LLM-based models use autoregressive decoding, which scales poorly with sequence length and introduces substantial inference latency, whereas CTC decoding operates in a single forward pass. Our models therefore offer a more favourable trade-off between ASR accuracy and inference cost, which is particularly relevant when the computational budget is limited.

% \vspace{0.1cm}
% \noindent
% \textbf{Summary.} Our approach with multilingual SFT for Ethiopian languages consistently outperform all baselines and achieve competitive performance relative to much larger OmniASR models, while using a fraction of the parameters. 
% These results demonstrate that targeted fine-tuning on in-domain data is an effective strategy for low-resource Ethiopian language ASR, and that strong recognition performance does not require autoregressive decoding with a language model.

\begin{table}[t]
\centering
\caption{WER (\%, lower is better) on FLEURS. Note that FLEURS data was included in the pre-training or fine-tuning data of all baseline models but was not seen by our models during training, making this a zero-shot out-of-domain evaluation for Ethio-ASR.}
\vspace{-6pt} 
\label{tab:wer-fleurs}
\begin{tabular}{lcc}
\toprule
\textbf{Model} & {\texttt{AMH}} & \texttt{ORM} \\
\midrule

\rowcolor{metaai}
\multicolumn{3}{l}{\textbf{MMS}} \\
\quad mms-1b-all            & 29.78 & 64.71 \\
\cmidrule{1-3}

\rowcolor{omni}
\multicolumn{3}{l}{\textbf{OmniASR}} \\
\quad  omniASR-ctc-300m-v2   & 34.41 & 77.53 \\
\quad  omniASR-ctc-1b-v2   & 48.23 & 68.38 \\
\quad  omniASR-ctc-3b-v2   & 20.59 & 62.76 \\
\quad  omniASR-ctc-7b-v2   & \text{16.19} & \text{61.96} \\
\cmidrule{1-3}

\quad  omniASR-llm-300m-v2   & 18.84 & 61.48 \\
\quad  omniASR-llm-1b-v2   & 19.97 & 58.11 \\
\quad  omniASR-llm-3b-v2   &    13.84	    &   56.98     \\
\quad omniASR-llm-7b-v2   &     \textbf{12.77}   &   \textbf{50.08}     \\
\midrule

\rowcolor{ourmodels}
\multicolumn{3}{l}{\textbf{This Work}} \\
\quad  Ethio-ASR \small{(\texttt{afrihubert})}  & 28.96 & 73.19 \\
\quad Ethio-ASR \small{(\texttt{mms-300m})}     & 29.21 & 72.39 \\
\quad  Ethio-ASR \small{(\texttt{mms-1b})}      & 23.05 & 73.52 \\
\quad Ethio-ASR \small{(\texttt{w2v-bert-2.0})}    & 19.17 & 70.47 \\

\bottomrule
\end{tabular}
\vspace{-12pt}
\end{table}

\vspace{0.1cm}
\noindent
\textbf{Multilingual vs.\ monolingual.} Training a separate \texttt{w2v-bert-2.0} model per language yields an average WER of 30.00\%, only 0.48 percentage points better than the single multilingual model (30.48\%). Since each language has  \textasciitilde190 hours of training data, we hypothesize that the multilingual model receives sufficient signal per language to learn effectively without relying on cross-lingual transfer. 
The negligible performance gap demonstrates that a single multilingual model can match dedicated language-specific models, which is practically significant in linguistically diverse societies like Ethiopia where deploying and maintaining a separate model for each spoken language is neither scalable nor practical.
% \vspace{0.1cm}
% \noindent
% \textbf{Monolingual upper bound.} The monolingual \texttt{SFT (w2v-bert-2.0)} models, where a single model was trained separately on each language, achieve an average WER of 30.00\%, only marginally better than the multilingual model (30.48\%). 
% This small gap suggests that our multilingual fine-tuning strategy achieves near-monolingual performance while requiring only a single model, which is practically significant in low-resource settings.

\vspace{0.1cm}
\noindent
\textbf{Summary.} Our  Ethio-ASR models developed via multilingual SFT outperform all baselines including the largest OmniASR model overall, while using a fraction of the parameters and with lower inference cost. These results demonstrate that targeted fine-tuning on in-domain data is an effective strategy for Ethiopian language ASR, and that strong recognition performance does not require large-scale autoregressive decoding.

\subsection{Evaluation on FLEURS Dataset} 

Table~\ref{tab:wer-fleurs} reports WER on the FLEURS benchmark \cite{conneau2023fleurs}. 
Unlike \texttt{mms-1b-all} and OmniASR, which included FLEURS data in their training, our Ethio-ASR models were never exposed to this data, making this strictly an out-of-domain evaluation. 
Despite this limitation, Ethio-ASR \texttt{(w2v-bert-2.0)} achieves a WER of 19.17\% on Amharic, approaching the performance of many OmniASR variants trained on FLEURS, suggesting that our models generalize reasonably well beyond their training domain.

\begin{table}[t]
\centering
\caption{Language identification (LID) in accuracy (\%). }
\label{tab:lid-acc}
\vspace{-6pt}
\begin{tabular}{lcc}
\toprule
\rowcolor{gray!15}
\color{Blue}\textbf{Model} & \color{Blue}\textbf{\# params} & \color{Blue}\textbf{Acc.} \\
\midrule

Ethio-ASR \small{(\texttt{afrihubert})}  & 94M & 99.92  \\
Ethio-ASR \small{(\texttt{mms-300m})}   & 300M & 99.92 \\
Ethio-ASR \small{(\texttt{mms-1b})}   &  1B & 99.91 \\
Ethio-ASR \small{(\texttt{w2v-bert-2.0})} &  600M  & 99.83  \\

\bottomrule
\end{tabular}
\vspace{-4pt}
\end{table}

\begin{table}[t]
\centering

\caption{WER (\%) under +LID and --LID conditions on the WAXAL test set. Values shown as mean $\pm$ half-width of the 95\% bootstrap CI. No difference is statistically significant ($p > 0.45$).}
\vspace{-6pt}
\label{tab:lid_comparison}
\begin{tabular}{lcc}
\toprule
\rowcolor{gray!15}
\color{Blue}\textbf{Model} & \color{Blue}\textbf{+LID} & \color{Blue}\textbf{--LID} \\
\midrule
Ethio-ASR \small{(\texttt{afrihubert})} & 37.93\ci{$\pm$1.32} & 38.04\ci{$\pm$1.23} \\ \\
Ethio-ASR \small{(\texttt{mms-300m})}   & 36.77\ci{$\pm$1.21} & 36.15\ci{$\pm$1.20} \\ \\
Ethio-ASR \small{(\texttt{mms-1b})}    & 33.94\ci{$\pm$1.24} & 33.86\ci{$\pm$1.31} \\ \\
Ethio-ASR \small{(\texttt{w2v-bert-2.0})}  & 33.02\ci{$\pm$1.28} & 33.28\ci{$\pm$1.28} \\ \\
\bottomrule
\end{tabular}
\vspace{-8pt}
\end{table}

The unexpectedly high WER on Oromo prompted a closer inspection of the data. 
We conducted a human evaluation in which native speakers were presented with audio samples alongside both the FLEURS reference transcription and our model's prediction (hypothesis from Ethio-ASR \texttt{w2v-bert-2.0}), and asked to judge which was more accurate. For Oromo, speakers rated 36.1\% of audio recordings as unintelligible, providing direct evidence that a substantial portion of the FLEURS Oromo data is noisy and that the high WER reflects recording-quality issues. 
For Amharic, speakers preferred the FLEURS reference in 55.1\% of cases, whereas our model's transcription was preferred in 43.4\% of cases, indicating reference transcription quality issues. 
% Although these issues are less severe, they are also present in the Amharic portion of FLEURS.
These findings are consistent with recent audits of FLEURS that have documented serious data quality issues in FLEURS~\cite{alabi25_interspeech, lau-etal-2025-data}.

% The unexpectedly high WER on Oromo, however, prompted a closer inspection of the data. 
% As we discuss in the following section, human evaluation revealed that a substantial proportion of Oromo samples in FLEURS are noisy or unintelligible even to native speakers, which likely inflates the WER independently of model quality.

\subsection{Language Identification Evaluation}

Table~\ref{tab:lid-acc} reports LID accuracy for all four Ethio-ASR models. All models achieve near-perfect language identification, with accuracies above 99.8\% regardless of encoder size or architecture, demonstrating that the CTC training objective is sufficient for reliable language identification.
%and its associated script without any dedicated LID module.

\section{Model Analysis}
\label{sec:analysis}
\subsection{Language Identification Ablation}

Our multilingual training setup prepends a LID token (i.e.,  \texttt{[LANG]}) to the text transcript. 
Here, we analyze whether or not the LID token improves the main ASR task by training multilingual models without it.  
Table~\ref{tab:lid_comparison} compares the effect of including the LID token (i.e.,  \texttt{[LANG]}) in the CTC target sequence during training. 
Across all pre-trained encoders, the (micro-averaged) WER differences between the +LID (with LID token) and --LID (without LID token)  conditions are negligible (all below 0.7\% absolute), with largely overlapping confidence intervals (CIs). 
A paired bootstrap significance test (with $n=1000$) confirms that none of the differences reach statistical significance ($p > 0.45$ in all cases). 
This indicates that jointly predicting a language token neither degrades transcription quality nor improves ASR performance, which is consistent with prior findings that multilingual end-to-end models implicitly learn to identify the input language \cite{omnilingualasrteam2025omnilingualasropensourcemultilingual}. Nevertheless, we release all models with the LID token enabled, as LID has clear practical utility in multilingual deployment scenarios.

\begin{table}[b!]
\centering
\small
\caption{WER (\%) by gender across languages (excluding Wolaytta). $\Delta$ = Male $-$ Female; positive values indicate higher error for male speakers.}
\label{tab:gender_comparison}
\begin{tabular}{lrrrr}
\toprule
 & \texttt{AMH} & \texttt{ORM} & \texttt{SID} & \texttt{TIR}  \\
\midrule
\rowcolor{gray!15}
\multicolumn{5}{l}{\color{Blue}\textbf{Ethio-ASR (\texttt{afrihubert})}} \\
\quad Male & $\qquad$  32.08 & 28.34 & 29.96 & 56.78  \\
\quad Female & 30.02 & 25.11 & 33.57 & 38.24  \\
\quad $\Delta$ & +2.06 & +3.23 & $-$3.61 & +18.54  \\
\midrule
\rowcolor{gray!15}
\multicolumn{5}{l}{\color{Blue}\textbf{Ethio-ASR (\texttt{mms-300m})}} \\
\quad Male & 32.81 & 29.17 & 31.63 & 58.14 \\
\quad Female & 30.86 & 26.64 & 34.77 & 39.40  \\
\quad $\Delta$ & +1.95 & +2.53 & $-$3.14 & +18.74  \\
\midrule
\rowcolor{gray!15}
\multicolumn{5}{l}{\color{Blue}\textbf{Ethio-ASR (\texttt{mms-1b})}} \\
\quad Male & 28.55 & 25.94 & 27.72 & 52.96  \\
\quad Female & 25.66 & 22.18 & 32.27 & 34.76 \\
\quad $\Delta$ & +2.89 & +3.76 & $-$4.55 & +18.20 \\

\midrule
\rowcolor{gray!15}
\multicolumn{5}{l}{\color{Blue}\textbf{Ethio-ASR (\texttt{w2v-bert-2.0})}}\\
\quad Male & 25.86 & 25.53 & 30.28 & 50.62 \\
\quad Female & 22.05 & 23.85 & 31.67 & 31.98\\
\quad $\Delta$ & +3.81 & +1.68 & $-$1.39 & +18.64  \\
\bottomrule
\end{tabular}
\end{table}

\begin{figure*}[t!]
    \centering
    \includegraphics[width=0.90\textwidth]{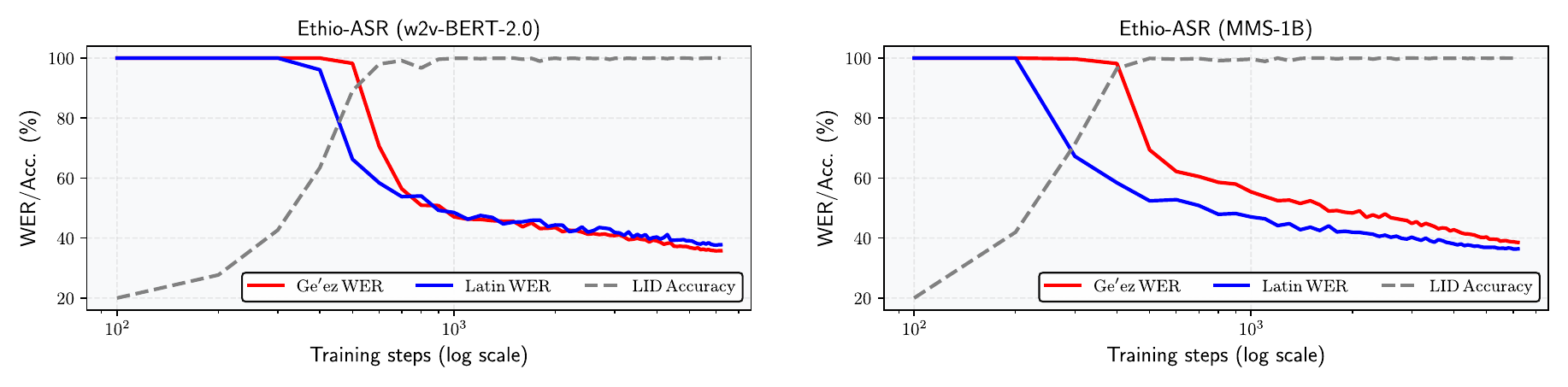}
    \vspace{-8pt}
    \caption{Training dynamics of Ethio-ASR with \texttt{w2v-BERT-2.0} and \texttt{MMS-1B} as pre-trained models. This analysis reveals that both models learn to transcribe Latin-based scripts before Ge'ez script. }
    \label{fig:learning_dynamics}

    \vspace{-8pt}
\end{figure*}

%, as evidenced by Latin WER dropping sharply in early training phase. Language identification saturates after 4-5\% of the training steps.
    
\subsection{Analyzing Gender Bias}
ASR systems usually exhibit performance disparities across demographic groups~\cite{feng2021quantifyingbiasautomaticspeech, martin2023bias, ngueajio2022hey}. 
To assess whether our models are affected by the gender imbalance present in the WAXAL training data (Table~\ref{tab:data_stats}), we report WER separately for male and female speakers across languages. 
We exclude Wolaytta from this analysis due to the lack of male speakers in the test split.
In Table~\ref{tab:gender_comparison}, we observe a consistent gender gap across models, most pronounced in Tigrinya, where male WER exceeds female WER by ~18\% absolute. 
Amharic and Oromo show smaller gaps (1--4\%) in the same direction. 
Sidaama is the exception, with female WER slightly higher than male across all models.
This preference for female speakers is likely driven by the imbalanced training data (Table~\ref{tab:data_stats}).

\subsection{Probing the Training Dynamics}

In the previous sections, we evaluated the ASR and LID performance of our models against strong baselines. 
Here, we ask a different question: how do these abilities evolve during training? 
To answer this question, we track the evolution of three skills the multilingual model must acquire: language identification, Latin-script transcription, and Ge\textquoteleft ez-script transcription. 
To do so, we fine-tune \texttt{w2v-BERT-2.0} and \texttt{MMS-1B} for a single epoch over all training samples (\textasciitilde {6.2$k$} steps), while evaluating on the validation split every 100 steps.

Figure~\ref{fig:learning_dynamics} shows that both models exhibit three learning phases. 
In the first phase (up to \textasciitilde4--5\% of training steps), LID accuracy rises steeply from chance-level performance toward saturation while WER remains high, indicating that the model first learns to identify the language before producing any meaningful transcription output. 
In the second phase (\textasciitilde5--10\% of training steps), WER drops rapidly for both scripts once LID has stabilized. 
%Latin WER declines slightly ahead of Ge\textquoteleft ez WER. 
%consistent with the greater difficulty of mapping acoustic-phonetic units to the larger and sparser Ge\textquoteleft ez grapheme inventory. 
% The two models differ in this phase: \texttt{MMS-1B} shows Latin WER declining earlier and more steeply than Ge\textquoteleft ez WER, which suggests that \texttt{MMS-1B} have better prior representations for Latin-script languages (i.e., Oromo, Sidaama, and Wolaytta). 
The two models differ in this phase: for \texttt{MMS-1B}, Latin WER declines earlier and more steeply than Ge\textquoteleft ez WER, suggesting that this model has stronger pre-trained representations for Latin-script languages (i.e., Oromo, Sidaama, and Wolaytta), whereas in \texttt{w2v-BERT-2.0} the two curves are more similar. 
In the third phase (beyond 10\% of training steps), both WER curves continue to decline gradually until convergence.
A qualitative analysis of a few validation samples showed that the models only apply fine-grained refinements to the transcriptions during the third phase.
% Both models converge smoothly with no signs of instability or overfitting within the single epoch.

These findings suggest that multilingual ASR models first learn to discriminate between input languages before learning to transcribe. 
The earlier decline of Latin WER relative to Ge\textquoteleft ez WER is consistent with the larger and sparser Ge\textquoteleft ez grapheme inventory discussed in Section~\ref{sec:asr_challenges}, though we cannot rule out that it also reflects greater exposure to Latin-script languages in our setup (i.e., 60.9\% of the training samples have Latin-based transcriptions).

\begin{figure*}[ht]
    \centering
    \includegraphics[width=0.95\textwidth]{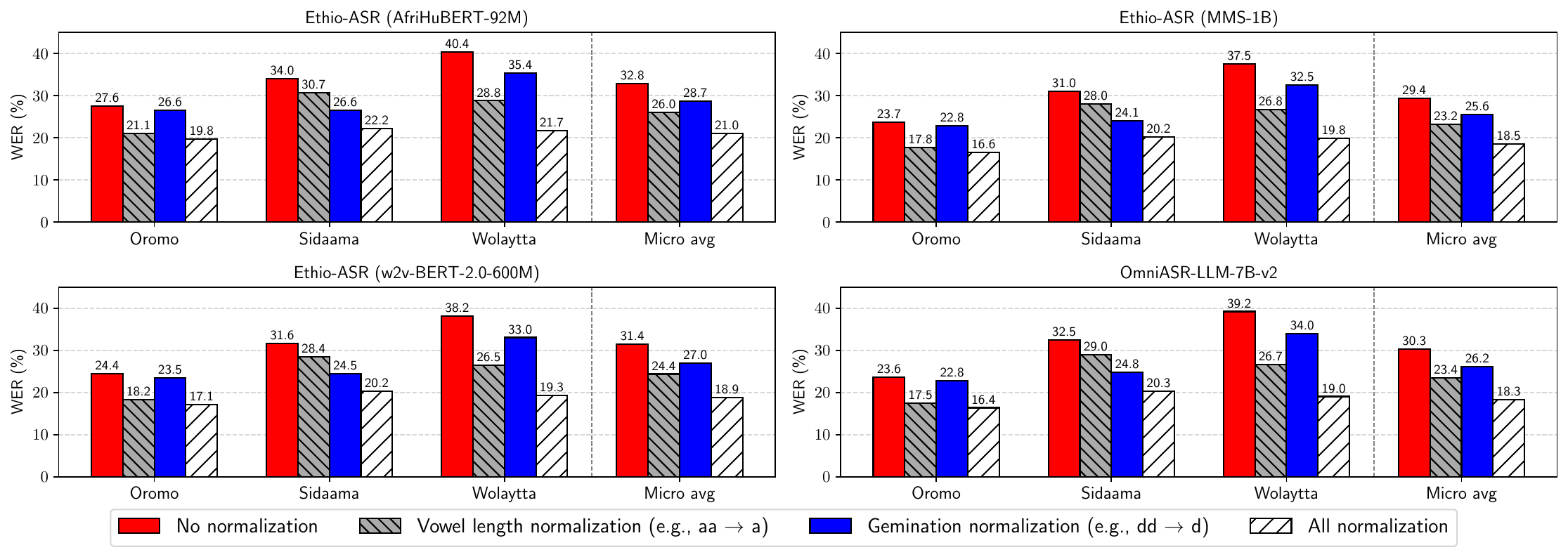}
    \vspace{-6pt}
    \caption{Effect of vowel length and geminate normalization on WER across Oromo, Sidaama, and Wolaytta. Results are shown for four conditions: no normalization, vowel length normalization, geminate normalization, and both normalized.}
    \label{fig:wer_reduction}
    \vspace{-12pt}
\end{figure*}

\subsection{Effect of Vowel Length and Gemination}
\label{sec:length_analysis}
As discussed in section \ref{sec:asr_challenges}, vowel length and consonant gemination are phonemically contrastive and are orthographically marked through letter doubling in Oromo, Sidaama, and Wolaytta. To quantify the contribution of these features to ASR errors, we apply post-hoc normalization to both the reference and model transcriptions before computing WER: vowel length normalisation collapses long vowels to their short counterparts (e.g., {\ipafont aa} $\rightarrow$ {\ipafont a}), and gemination normalisation reduces geminate consonants to singletons (e.g., {\ipafont dd} $\rightarrow$ {\ipafont d}). 
We evaluate four conditions: no normalization, vowel length normalization only, gemination normalization only, and both combined. It is important to note that these normalizations are applied purely as an analytical tool to isolate the contribution of each feature to WER; they do not constitute a valid preprocessing step for a deployed ASR system or before model training, since they collapse lexically distinct word tokens. 

Figure~\ref{fig:wer_reduction} shows the results across four models. 
We observe several consistent patterns. 
First, vowel length is the dominant source of errors in Oromo and Wolaytta, while consonant gemination is more prominent in Sidaama. 
Second, combined normalization yields the largest gains across all models, with micro-average WER dropping by \textasciitilde10--13 percentage points, representing a relative reduction of around 35--40\%. 
Note that this pattern holds across all four models, including \texttt{OmniASR-LLM-7B}, which is substantially larger than our models and uses autoregressive decoding rather than CTC.

These results provide empirical evidence that vowel length and gemination are substantial sources of ASR errors in Ethiopian languages, complementing our theoretical discussion in Section~\ref{sec:asr_challenges}. 
The consistency of these patterns across models of varying size and decoding strategies suggests that these errors cannot be addressed by model scaling or autoregressive decoding alone, but rather require modeling innovations explicitly designed to handle prominent linguistic contrasts in the languages.

\section{Discussion}
The results presented in this paper point to several broader observations about multilingual ASR for low-resource languages. 
First, our findings confirm that supervised fine-tuning on target-language data remains a highly effective strategy, even compared with models that are an order of magnitude larger. 
The strong performance of our CTC-based models relative to OmniASR LLM variants suggests that, for languages with sufficient, high-quality training data, architectural complexity and scale are not the primary bottlenecks. Second, the linguistic analyses reveal that a substantial portion of the remaining ASR errors can be attributed to specific phonological features, namely vowel length and consonant gemination, that are not well captured by current end-to-end adaptation strategies, regardless of model size or decoding approach. 
This suggests that future progress will require not only more data but also models explicitly designed to account for the linguistic features of the targeted languages.
Finally, the gender bias analysis highlights that training data imbalance directly translates into performance disparities, most severely in Tigrinya. 
This is a data collection issue rather than a modeling one, and highlights the importance of demographically balanced corpus design for low-resource languages where re-collection is costly.

\section{Related Work}
\label{sec:related}

% ASR for Ethiopian languages have seen a gradual transition from traditional statistical modeling frameworks to deep neural network-based architectures. 
% Early work primarily focused on Amharic, with later expansion to Oromo, Tigrigna, and Wolaytta as additional speech resources became available. 
% The primary hurdle for the development of Ethiopian ASR has always been the scarcity of large-scale transcribed speech corpora, coupled with by the high morphological complexity of these languages \cite{TACHBELIE2014181, tachbelie-etal-2020-analysis}. 

% Large-vocabulary read speech corpora for Amharic, Tigrinya, Oromo, and Wolaytta were developed by Addis Ababa University (AAU), each containing \textasciitilde22 hours of speech data \cite{abate-etal-2020-large-vocabulary}, which have  enabled the development of baseline ASR systems. 
% % These corpora enabled the development of baseline ASR systems achieving WERs of 37.65\% (Amharic), 31.03\% (Tigrigna), 38.02\% (Oromo), and 33.89\% (Wolaytta) \cite{abate-etal-2020-large-vocabulary}. 
% For Oromo, the recently introduced Sagalee dataset provides 100 hours of crowdsourced speech data collected across diverse environments, substantially expanding available resources \cite{abu2025sagalee}. 
% \section{Related Work}

% \label{sec:related}

ASR for Ethiopian languages has seen a gradual transition from traditional statistical modeling frameworks to deep neural network-based architectures, with early work focused predominantly on Amharic and later expansion to Oromo, Tigrinya, and Wolaytta as additional speech resources became available.

\vspace{0.1cm}
\noindent
\textbf{Speech Resources.} %The primary challenge has been the scarcity of large-scale transcribed speech corpora, coupled with the high morphological complexity of Ethiopian languages~\cite{TACHBELIE2014181, tachbelie-etal-2020-analysis}. 
%Large-vocabulary read speech corpora for Amharic, Tigrinya, Oromo, and Wolaytta were developed by Addis Ababa University (AAU), each containing \textasciitilde22 hours of speech~\cite{abate-etal-2020-large-vocabulary}. 
%For Oromo, the more recently introduced Sagalee dataset provides 100 hours of crowdsourced speech collected across diverse environments~\cite{abu2025sagalee}. 
%Cross-lingual transfer strategies have also been explored using the GlobalPhone corpus~\cite{tachbelie-etal-2020-analysis}, though this resource is distributed under a paid research license via ELRA and is not freely accessible.
% Cross-lingual strategies have also been explored using the GlobalPhone corpus \cite{tachbelie-etal-2020-analysis}. 
% Unfortunatel, however, the GlobalPhone corpus is not public and requires the purchase of a license. 
% Phonetic overlap analysis between Ethiopian languages and GlobalPhone languages such as Turkish and Croatian has been used to inform multilingual ASR modeling. These findings confirm that phonetic relatedness across language families can support transfer-based approaches.
The primary challenge in speech research for African languages, including Ethiopian languages, is the scarcity of large-scale, high-quality transcribed speech corpora, in addition to the morphological complexity of these languages ~\cite{TACHBELIE2014181, tachbelie-etal-2020-analysis,alabi-etal-2025-charting}. 
Nevertheless, a number of speech resources have been developed for Ethiopian languages, including: an over-20-hour Amharic speech corpus created via crowdsourcing~\cite{TACHBELIE2014181}; a read-speech corpus covering four languages, Amharic, Tigrigna, Oromo, and Wolaytta~\cite{abate-etal-2020-large-vocabulary}, with over 22 hours of speech per language; and a 100-hour crowdsourced speech corpus for the Oromo Sagalee dialect~\cite{abu2025sagalee}. In addition, the FLEURS corpus~\cite{conneau2023fleurs}, a multilingual resource covering over 100 languages, includes Amharic and Oromo. 
The most recent large speech resource is WAXAL~\cite{diack2026waxal}, covering five Ethiopian languages, and it has been used in this study.
%Large-vocabulary read speech corpora for Amharic, Tigrinya, Oromo, and Wolaytta were developed by Addis Ababa University (AAU), each containing \textasciitilde22 hours of speech~\cite{abate-etal-2020-large-vocabulary}. 
%For Oromo, the more recently introduced Sagalee dataset provides 100 hours of crowdsourced speech collected across diverse environments~\cite{abu2025sagalee}. 
% Cross-lingual transfer strategies have also been explored using the GlobalPhone corpus~\cite{tachbelie-etal-2020-analysis}, though this resource is distributed under a paid research license via ELRA and is not freely accessible.

\vspace{0.1cm}
\noindent
\textbf{ASR Modeling Approaches.} Early efforts at ASR modeling for Ethiopian languages employed GMM-HMM frameworks with word-based and morpheme-based lexical modeling~\cite{TACHBELIE2014181}, which required extensive feature engineering. More recent studies have shifted toward end-to-end neural architectures, including deep neural network (DNN)-based acoustic models~\cite{9053883}, fully end-to-end DNN architectures~\cite{9415020, info12020062}, and transformer-based ASR systems~\cite{abu2025sagalee,Adnew2024SemanticallyCA}. The development of multilingual resources, such as FLEURS and WAXAL, has facilitated these neural approaches by providing larger and more diverse training data, enabling pretraining and cross-lingual transfer learning~\cite{alabi25_interspeech,pratap2024mms,barrault2023seamless}. Overall, recent work in Ethiopian language ASR has increasingly focused on data-efficient, end-to-end neural architectures that leverage pretrained encoders and cross-lingual transfer learning to enhance performance in low-resource settings.

\vspace{0.1cm}
\noindent
\textbf{Multilingual Approaches.} Multilingual ASR has been proposed to address data scarcity. 
It has been demonstrated that a DNN-based Oromo acoustic model could be used to recognize Wolaytta speech, achieving 48.34\% WER without Wolaytta training data \cite{tachbelie-etal-2020-dnn}. 
%and significantly reducing error when only 30 minutes of Wolaytta speech were added \cite{tachbelie-etal-2020-dnn}, which supports the effectiveness of cross-lingual transfer between related languages.
Likewise, transfer learning from high-resource languages has shown strong improvements for Amharic. 
English and Mandarin acoustic models were adapted to Amharic, reducing WER from 38.72\% to 24.50\% in the best case \cite{woldemariam-2020-transfer}. 
The study further demonstrated that linguistic relatedness influences transfer across different languages. 
Recent research efforts have shifted toward end-to-end DNN-based architectures that leverage language-specific tokenizers \cite{9415020, info12020062}.

%\vspace{0.1cm}
%\noindent
%\textbf{End-to-End (E2E) Architectures.} Recent research efforts have shifted toward end-to-end DNN-based architectures that leverage language-specific tokenizers \cite{9415020, info12020062}. 
% Abate et al. \cite{9415020} explored end-to-end multilingual ASR for four Ethiopian languages. 
% Emiru et al. \cite{info12020062} proposed a hybrid CTC-attention model with phoneme-based subword units and byte-pair encoding.
%, achieving significant WER reductions over character-based baselines. 
%More recent studies investigate zero-shot and transliteration-based transfer methods for Amharic ASR \cite{nigatu-aldarmaki-2025-exploring}, further demonstrating that cross-lingual transfer remains a critical direction for low-resource Ethiopian languages. 
% Overall, Ethiopian ASR research has evolved from statistical GMM-HMM systems to hybrid DNN-HMM models and, more recently, to fully end-to-end neural architectures leveraging multilingual and transfer learning strategies.

\section{Conclusion}
In this paper, we presented Ethio-ASR, a suite of multilingual CTC-based models for five Ethiopian languages that jointly perform ASR and language identification. Our models outperform all existing systems, including the largest OmniASR variant, while using a fraction of the parameters and inference cost. All models, code, and evaluation resources will be released upon publication to support reproducibility and community-driven development of speech technology for Ethiopian and other underrepresented languages.

% \section{Generative AI Use Disclosure}

% Generative AI tools were used exclusively for editing and polishing assistance, coding support for training and evaluation modules, as well as code debugging, and manuscript proofreading, including grammar and typographical corrections. 
% No part of the scientific content, experimental design, analysis, citations, or conclusions was generated by AI tools. All authors have read, approved, and take full responsibility for the content of this paper.

\vspace{0.2cm}
\noindent
\textbf{Acknowledgments.} The authors gratefully acknowledge Howard Lakougna (Gates Foundation) and Polly Harlow (CLEAR Global) for their encouragement to pursue this direction and for insightful discussions in the early stages of this work. 
Badr M. Abdullah and Jesujoba O. Alabi are funded by the Deutsche Forschungsgemeinschaft (DFG, German Research Foundation) – Project-ID 232722074 – SFB 1102. 
Israel Abebe Azime is funded by the German Federal Ministry of Education and Research and the German federal states (\url{http://www.nhr-verein.de/en/our-partners}) as part of the National High-Performance Computing (NHR) joint funding program.

\end{document}